\documentclass{article}

\usepackage[dblblindworkshop, final]{neurips_2025}
\usepackage{hyperref}
\hypersetup{
  pdfauthor={Sayash Raaj Hiraou},
  pdftitle={Context-Masked Meta-Prompting for Privacy-Preserving LLM Adaptation in Finance},
  pdfsubject={Generative AI in Finance},
  pdfkeywords={LLM, Finance, Privacy, Meta-Prompting, Prompts}
}

\usepackage[utf8]{inputenc}
\usepackage[T1]{fontenc}
\usepackage{booktabs}
\usepackage{amsfonts}
\usepackage{nicefrac}
\usepackage{microtype}
\usepackage{xcolor}
\usepackage{graphicx}
\usepackage{subcaption}
\usepackage{float}
\usepackage{enumitem}

\title{Context-Masked Meta-Prompting for Privacy-Preserving LLM Adaptation in Finance}

\author{%
  Sayash Raaj Hiraou \\ 
  Fidelity Investments \\ 
  Bengaluru, India \\ 
  \texttt{sayashraaj@alumni.iitm.ac.in} \\ 
}
\workshoptitle{Evaluating the Evolving LLM Lifecycle}

\begin{document}

\maketitle

\begin{abstract}
The increasing reliance on Large Language Models (LLMs) in sensitive domains like finance necessitates robust methods for privacy preservation and regulatory compliance. This paper presents an iterative meta-prompting methodology designed to optimise hard prompts without exposing proprietary or confidential context to the LLM. Through a novel regeneration process involving feeder and propagation methods, we demonstrate significant improvements in prompt efficacy. Evaluated on public datasets serving as proxies for financial tasks such as SQuAD for extractive financial Q\&A, CNN/DailyMail for news summarisation, and SAMSum for client interaction summarisation, our approach, utilising GPT-3.5 Turbo, achieved a 103.87\% improvement in ROUGE-L F1 for question answering. This work highlights a practical, low-cost strategy for adapting LLMs to financial applications while upholding critical privacy and auditability standards, offering a compelling case for its relevance in the evolving landscape of generative AI in finance.
\end{abstract}

\section{Introduction}

The financial industry is exploring Large Language Models (LLMs) for tasks such as compliance Q\&A, research summarisation, and automated risk assessment. However, strict regulations (e.g., GDPR, SEC guidelines) and internal governance prohibit exposing client data, proprietary models, or internal research to external systems. This rules out many common adaptation approaches that require sharing task context.

The challenge is therefore not just a natural language processing (NLP) problem but a financial integration problem: \textit{how to tailor LLMs to domain needs without breaching confidentiality or auditability?}. We address this with a \textbf{context-masked meta-prompting framework} that refines human-readable "hard" prompts \cite{wen2023hard,qin2021learning} through an LLM-as-optimiser process \cite{yang2024large,ma2024large}, while ensuring all sensitive data remains within a secure perimeter.

Evaluated on public datasets as proxies for financial NLP tasks, our approach delivers substantial performance gains using cost-efficient models, aligning LLM optimisation with the operational and regulatory realities of finance.

\section{Related Work}

Generative AI is increasingly applied in finance for tasks such as compliance checks, market news summarisation, and client-–advisor interaction analysis. These use cases involve sensitive data — proprietary strategies, client records, or internal research — that cannot leave secure systems due to regulations like GDPR and strict internal governance. This makes direct fine-tuning or prompt-based adaptation of LLMs, even in few-shot settings \cite{brown2020language}, difficult to deploy.

Outside finance, lightweight adaptation techniques such as hard prompt optimisation \cite{wen2023hard}, mixtures of soft prompts \cite{qin2021learning}, automatic hint generation \cite{sun2023autohint}, and meta-prompting with LLMs as optimisers \cite{yang2024large,ma2024large} have shown strong task-specific gains. Other work explores self-referential prompt evolution \cite{fernando2023promptbreeder,gupta-etal-2024-metareflection} and structured prompt pattern catalogues \cite{white2023prompt}. However, these approaches generally assume the model can access task context — an assumption incompatible with high-compliance financial environments.

Our work adapts these ideas into a context-masked meta-prompting framework that enables LLM optimisation for finance-relevant tasks without exposing any sensitive data.

\section{Context-Masked Meta-Prompting Methodology}
\begin{figure}[h!]
    \centering
    \includegraphics[width=0.9\linewidth]{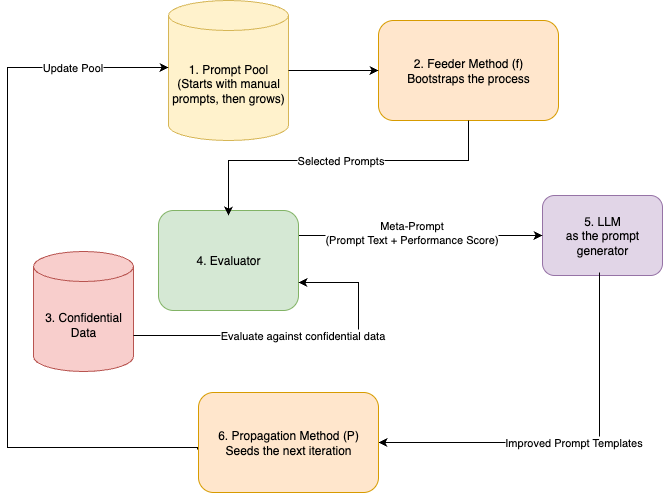}
    \caption{Conceptual overview of the context-masked meta-prompting loop}
    \label{fig:methodology}
\end{figure}

Our framework enables iterative prompt optimisation while respecting a strict context-masking principle. The optimisation process is driven by a meta-prompt that instructs an LLM to generate improved prompt templates \cite{sun2023autohint} based on the performance of previous ones, without ever seeing the confidential data used for performance evaluation. The entire process occurs within a secure internal system \cite{zhou2022prompt} that only sends sanitised, context-free data to the external LLM API, as illustrated in Figure \ref{fig:methodology}.

The core of our approach is a regeneration method, which consists of two components:
\begin{itemize}[topsep=0pt,itemsep=2pt,parsep=0pt,partopsep=0pt]
    \item \textbf{Feeder Methods ($f$):} These determine the initial sampling of prompts to bootstrap the process. We test two strategies: Feeder Method A ($f_a$), which samples only the top-$n$ best-performing manually written prompts, and Feeder Method B ($f_b$), which samples both the top-$n$ and bottom-$n$ prompts to provide the LLM with both positive and negative examples.
    \item \textbf{Propagation Methods ($P$):} These govern how prompts from one iteration are used to seed the next. We investigate Propagation Method A ($P_a$), which cumulatively concatenates all previously generated prompts, and Propagation Method B ($P_b$), which uses the feeder logic to resample from the growing pool of generated prompts, thereby managing context window size.
\end{itemize}
By combining these components, we evaluate four distinct pipeline strategies: $f_aP_a$ (Method A), $f_bP_a$ (Method B), $f_aP_b$ (Method C), and $f_bP_b$ (Method D). This systematic exploration allows for a granular analysis of different optimisation strategies impacting performance and prompt diversity.

\section{Experiments and Key Results}

\paragraph{Experimental Setup}
All experiments were conducted using \texttt{GPT-3.5 Turbo} (2024 release) with a temperature of 1.0 to encourage diversity. 
This model was chosen deliberately to test our method on a cost-efficient, widely accessible LLM, simulating a realistic setting for financial institutions with budget, latency, and compliance constraints.
We used established public datasets as finance-relevant proxies for common tasks:
\begin{itemize}
    \item \textbf{SQuAD} \cite{rajpurkar2016squad}: extractive financial Q\&A, analogous to retrieving figures from reports.
    \item \textbf{CNN/DailyMail} \cite{NIPS2015_afdec700}: summarising news and market analysis.
    \item \textbf{SAMSum} \cite{Gliwa_2019}: summarising client--advisor or compliance-related interactions.
\end{itemize}

Performance was measured by the ROUGE-L F1 score over 10 iterations, comparing against the baseline of manually written prompts ($S_m$). The initial set of prompts selected by the feeder method before the first generative iteration is termed $S_f$.

\begin{figure}[h]
    \centering
    \begin{subfigure}[b]{0.48\linewidth}
        \centering
        \includegraphics[width=\textwidth]{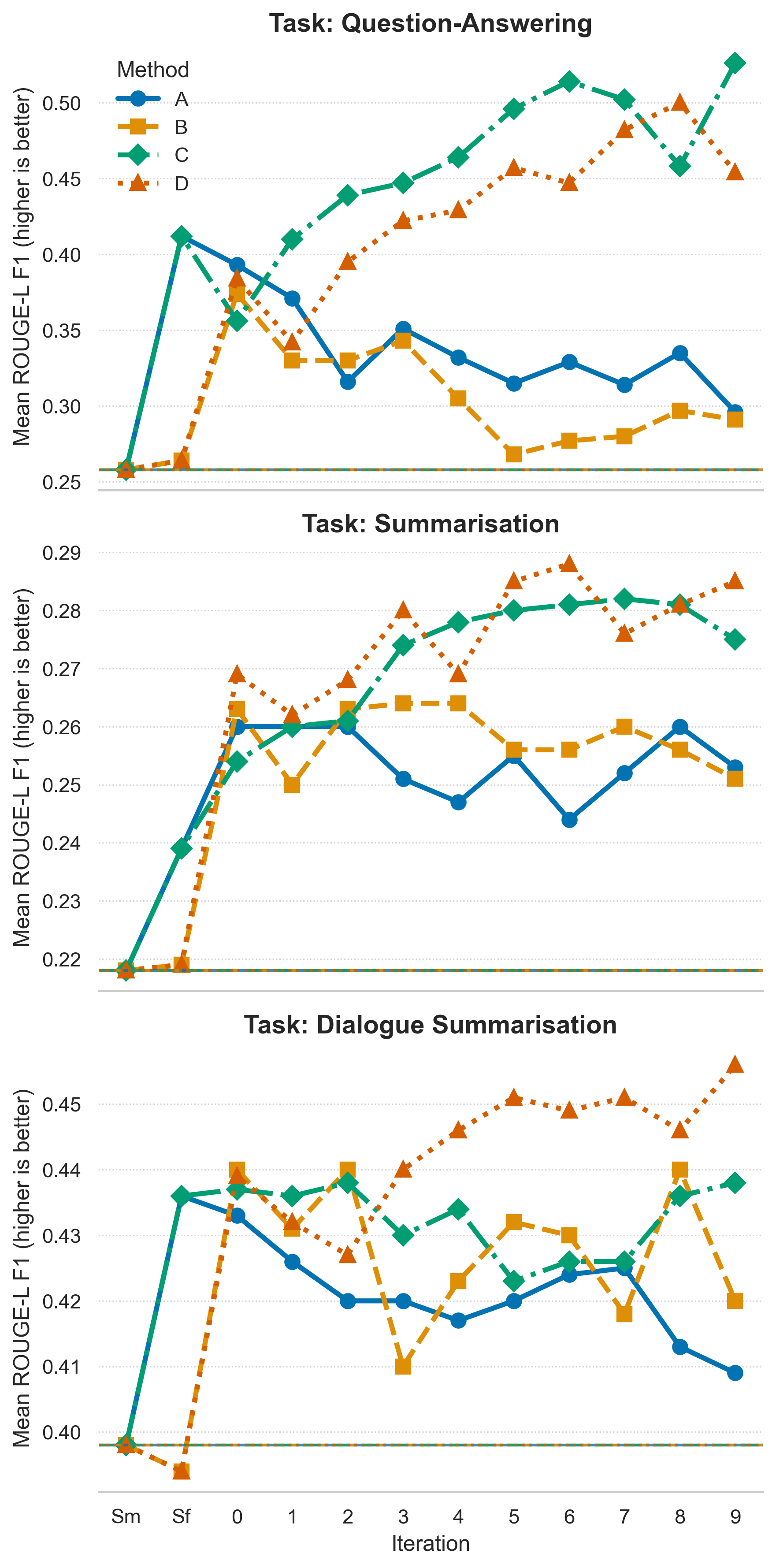}
        \caption{Mean ROUGE-L F1 Scores (higher is better)}
        \label{fig:mean_scores_sub}
    \end{subfigure}
    \hfill 
    \begin{subfigure}[b]{0.48\linewidth}
        \centering
        \includegraphics[width=\textwidth]{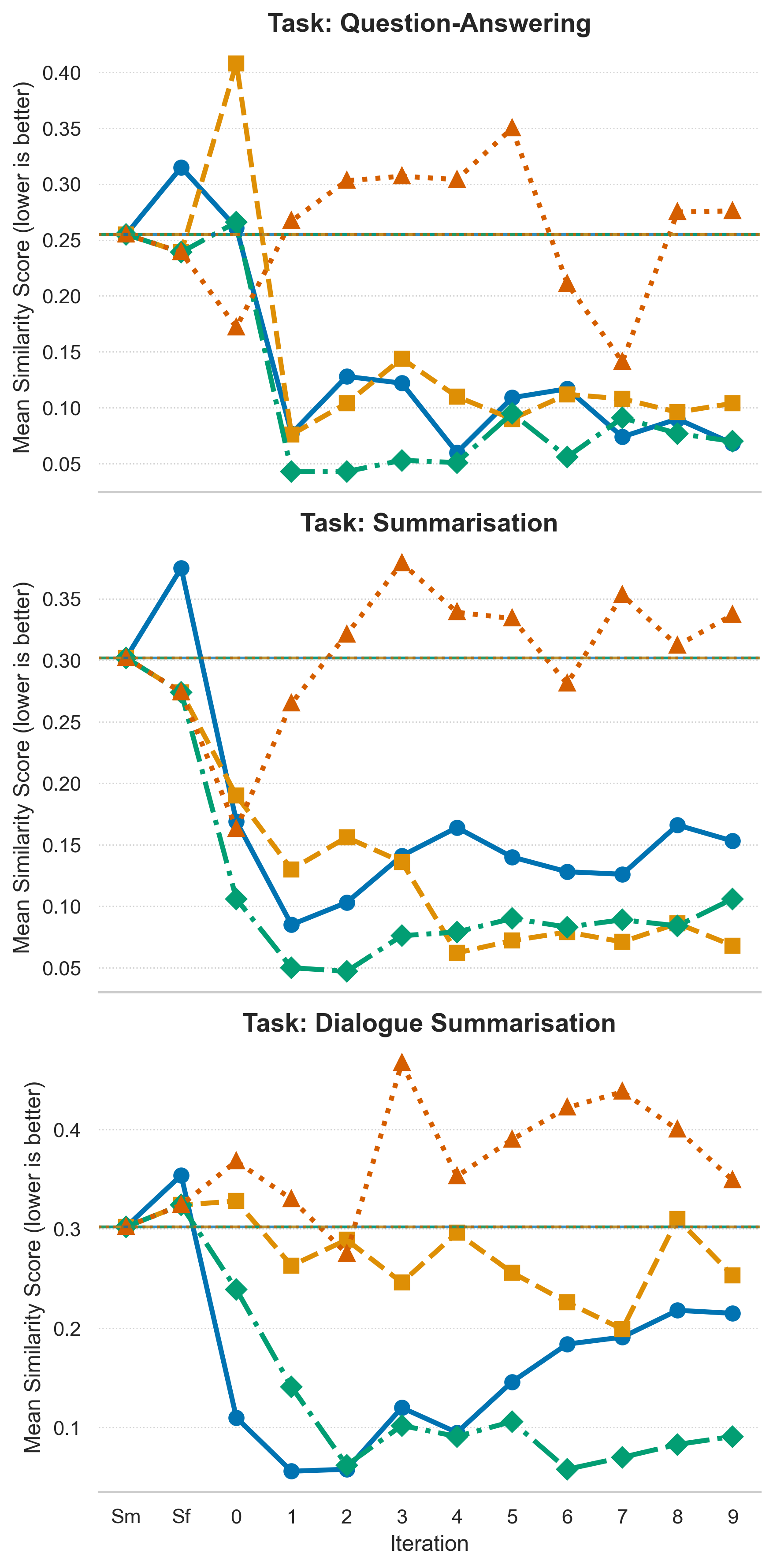}
        \caption{Prompt Similarity Scores (lower is better)}
        \label{fig:similarity_scores_sub}
    \end{subfigure}
    \caption{Performance and diversity of prompts over 10 iterations. (a) ROUGE-L scores show sustained performance gains with Method C ($f_aP_b$) being most effective. (b) Similarity scores show the evolution of prompt diversity, with Method C maintaining a good balance.}
    \label{fig:combined_results}
\end{figure}

\paragraph{Performance Improvement}
Our results show that iterative meta-prompting significantly improves performance over the baseline. As shown in Figure \ref{fig:combined_results}, methods using the resampling Propagation Method B ($P_b$) consistently outperformed those using the cumulative method ($P_a$), which suffered from context window limitations. The most striking result was on the Question-Answering task, where method $f_aP_b$ (Method C) achieved a mean ROUGE-L F1 score of 0.526 after 9 iterations, a 103.87\% improvement over the manual baseline score of 0.258. This demonstrates that our privacy-preserving technique can more than double the effectiveness of prompts for precise information extraction tasks crucial in finance.

Analysis of prompt diversity revealed that method $f_aP_b$ also provides a strong balance between high performance and the generation of varied prompts, a key factor for robust deployment. Full per-iteration scores and similarity analysis are provided in Appendix A.

\section{Financial Applications and Conclusion}

\paragraph{Practical Deployment and Applications}
A financial firm can adopt this framework by keeping all proprietary data within its secure perimeter. An internal service would evaluate prompts against this data, then send only the context-free prompt text and its performance score to an external LLM via a secure API. The optimised prompt templates returned by the LLM are then integrated back into internal applications. This architecture strictly maintains data confidentiality.
Key applications include:
\begin{itemize}
    \item \textbf{Private Compliance Q\&A:} Optimise prompts to answer questions against internal regulatory documents without exposing proprietary legal interpretations.
    \item \textbf{Proprietary Research Summarisation:} Create effective summarisation prompts for sensitive analyst reports without the reports ever leaving the firm's environment.
    \item \textbf{Auditable Risk Checkers:} Bootstrap and refine human-readable instruction templates for automated risk and fraud detection systems, ensuring transparency.
\end{itemize}

\paragraph{Limitations and Responsible Deployment}
While promising, this work has limitations. The methodology was validated on proxy datasets; future work should test it on financial data. Ethically, a key risk is that the meta-prompting process could amplify biases inherent in the LLM, whose training data is opaque. Responsible deployment therefore necessitates continuous bias auditing and robust human-in-the-loop governance for critical decisions. A comprehensive discussion of these points is available in Appendices B and C.

\paragraph{Conclusion}
This paper presented a context-masked meta-prompting framework that enables significant LLM performance gains while adhering to the stringent privacy and auditability requirements of the financial industry. By demonstrating a 103.87\% improvement in a key proxy task representative of financial NLP applications using a resource-efficient model, we have shown a practical, low-cost method for adapting LLMs to sensitive domains. This work provides a viable and responsible pathway for deploying effective and interpretable generative AI in finance.

\paragraph{Disclaimer}
The views expressed herein are those of the author and do not necessarily reflect the views of Fidelity Investments.

\bibliographystyle{unsrt}
\bibliography{custom}

\appendix
\newpage
\section{Supplementary Results and Details}
\subsection{Detailed Performance and Similarity Scores}
The following tables provide the exact numerical results for the mean ROUGE-L F1 scores (Table \ref{tab:mean_scores}) and similarity scores (Table \ref{tab:similarity_scores}) for all methods across all 10 iterations of the experiment. These tables form the basis for the analysis presented in the main paper.

\begin{table}[h!]
    \centering
    \caption{Detailed Mean ROUGE-L F1 Scores for All Tasks and Methods Across Iterations. This table preserves the exact numerical results from the original study.}
    \label{tab:mean_scores}
    \resizebox{\textwidth}{!}{%
    \begin{tabular}{lcccccccccccc}
        \toprule
        & \multicolumn{4}{c}{Question-Answering} & \multicolumn{4}{c}{Summarisation} & \multicolumn{4}{c}{Dialogue Summarisation} \\
        \cmidrule(lr){2-5} \cmidrule(lr){6-9} \cmidrule(lr){10-13}
        Method & A & B & C & D & A & B & C & D & A & B & C & D \\
        \midrule
        Sm & 0.258 & 0.258 & 0.258 & 0.258 & 0.218 & 0.218 & 0.218 & 0.218 & 0.398 & 0.398 & 0.398 & 0.398 \\
        Sf & 0.412 & 0.264 & 0.412 & 0.264 & 0.239 & 0.219 & 0.239 & 0.219 & 0.436 & 0.394 & 0.436 & 0.394 \\
        0 & 0.393 & 0.374 & 0.356 & 0.384 & 0.260 & 0.263 & 0.254 & 0.269 & 0.433 & 0.440 & 0.437 & 0.439 \\
        1 & 0.371 & 0.330 & 0.410 & 0.342 & 0.260 & 0.250 & 0.260 & 0.262 & 0.426 & 0.431 & 0.436 & 0.432 \\
        2 & 0.316 & 0.330 & 0.439 & 0.395 & 0.260 & 0.263 & 0.261 & 0.268 & 0.420 & 0.440 & 0.438 & 0.427 \\
        3 & 0.351 & 0.343 & 0.447 & 0.422 & 0.251 & 0.264 & 0.274 & 0.280 & 0.420 & 0.410 & 0.430 & 0.440 \\
        4 & 0.332 & 0.305 & 0.464 & 0.429 & 0.247 & 0.264 & 0.278 & 0.269 & 0.417 & 0.423 & 0.434 & 0.446 \\
        5 & 0.315 & 0.268 & 0.496 & 0.457 & 0.255 & 0.256 & 0.280 & 0.285 & 0.420 & 0.432 & 0.423 & 0.451 \\
        6 & 0.329 & 0.277 & 0.514 & 0.447 & 0.244 & 0.256 & 0.281 & 0.288 & 0.424 & 0.430 & 0.426 & 0.449 \\
        7 & 0.314 & 0.280 & 0.502 & 0.482 & 0.252 & 0.260 & 0.282 & 0.276 & 0.425 & 0.418 & 0.426 & 0.451 \\
        8 & 0.335 & 0.297 & 0.458 & 0.500 & 0.260 & 0.256 & 0.281 & 0.281 & 0.413 & 0.440 & 0.436 & 0.446 \\
        9 & 0.296 & 0.291 & 0.526 & 0.454 & 0.253 & 0.251 & 0.275 & 0.285 & 0.409 & 0.420 & 0.438 & 0.456 \\
        \bottomrule
    \end{tabular}
    }
\end{table}

\begin{table}[h!]
    \centering
    \caption{Detailed Similarity Scores for All Tasks and Methods Across Iterations. This table preserves the exact numerical results from the original study.}
    \label{tab:similarity_scores}
    \resizebox{\textwidth}{!}{%
    \begin{tabular}{lcccccccccccc}
        \toprule
        & \multicolumn{4}{c}{Question-Answering} & \multicolumn{4}{c}{Summarisation} & \multicolumn{4}{c}{Dialogue Summarisation} \\
        \cmidrule(lr){2-5} \cmidrule(lr){6-9} \cmidrule(lr){10-13}
        Method & A & B & C & D & A & B & C & D & A & B & C & D \\
        \midrule
        Sm & 0.255 & 0.255 & 0.255 & 0.255 & 0.302 & 0.302 & 0.302 & 0.302 & 0.302 & 0.302 & 0.302 & 0.302 \\
        Sf & 0.315 & 0.239 & 0.315 & 0.239 & 0.375 & 0.274 & 0.375 & 0.274 & 0.354 & 0.324 & 0.354 & 0.324 \\
        0 & 0.261 & 0.408 & 0.266 & 0.172 & 0.169 & 0.190 & 0.106 & 0.163 & 0.110 & 0.328 & 0.239 & 0.368 \\
        1 & 0.077 & 0.076 & 0.043 & 0.267 & 0.085 & 0.130 & 0.050 & 0.265 & 0.056 & 0.263 & 0.141 & 0.330 \\
        2 & 0.128 & 0.104 & 0.043 & 0.303 & 0.103 & 0.156 & 0.047 & 0.321 & 0.058 & 0.289 & 0.062 & 0.275 \\
        3 & 0.122 & 0.144 & 0.053 & 0.307 & 0.141 & 0.136 & 0.076 & 0.379 & 0.120 & 0.246 & 0.102 & 0.467 \\
        4 & 0.060 & 0.110 & 0.051 & 0.304 & 0.164 & 0.062 & 0.079 & 0.339 & 0.095 & 0.296 & 0.091 & 0.353 \\
        5 & 0.109 & 0.090 & 0.095 & 0.350 & 0.140 & 0.072 & 0.090 & 0.334 & 0.146 & 0.256 & 0.106 & 0.390 \\
        6 & 0.117 & 0.112 & 0.056 & 0.211 & 0.128 & 0.079 & 0.083 & 0.281 & 0.184 & 0.226 & 0.058 & 0.422 \\
        7 & 0.074 & 0.108 & 0.091 & 0.141 & 0.126 & 0.071 & 0.089 & 0.353 & 0.191 & 0.199 & 0.070 & 0.438 \\
        8 & 0.090 & 0.096 & 0.077 & 0.275 & 0.166 & 0.086 & 0.084 & 0.312 & 0.218 & 0.310 & 0.083 & 0.400 \\
        9 & 0.068 & 0.104 & 0.070 & 0.276 & 0.153 & 0.068 & 0.106 & 0.337 & 0.215 & 0.253 & 0.091 & 0.349 \\
        \bottomrule
    \end{tabular}
    }
\end{table}

\section{Limitations and Future Work}
Our framework demonstrates significant promise, but we acknowledge several limitations that present avenues for future work.
\begin{itemize}
    \item \textbf{Proxy Datasets:} Experiments were conducted on public NLP datasets as proxies. Future work should prioritise validation on anonymised or synthetic financial datasets to confirm efficacy in a direct financial context.
    \item \textbf{Single LLM:} All experiments used GPT-3.5 Turbo. Future research should include LLM ablations with other models (including open-source alternatives) to test the generalisability of the optimisation process.
    \item \textbf{Decision-Quality Metrics:} Evaluation relied on ROUGE-L. Future work could benefit from using downstream, task-specific financial metrics (e.g., accuracy of extracted financial data, portfolio signal quality) to measure practical value.
\end{itemize}

\section{Broader Impact \& Ethics}
The primary positive impact of this work is enabling privacy-preserving AI in finance, reducing data leakage risks and fostering trust. However, any effective optimisation technique carries risks. A key ethical consideration is bias amplification. The meta-prompting process could inadvertently reinforce biases present in either the initial prompts or the LLM itself, leading to skewed outputs in sensitive applications like credit assessment. To mitigate this, we strongly recommend that any deployment of this method be accompanied by:
\begin{itemize}
    \item \textbf{Rigorous Bias Auditing:} Continuous monitoring of both input and output prompts for demographic or other biases.
    \item \textbf{Human-in-the-Loop Governance:} Ensuring human oversight for all critical financial decisions derived from LLM outputs.
    \item \textbf{Full Auditability:} Maintaining transparent logs of the prompt evolution process to ensure that the logic driving the LLM remains interpretable and compliant.
\end{itemize}

\end{document}